\documentclass[sigconf]{aamas}  

\usepackage{booktabs}
\usepackage{adjustbox}
\usepackage{makecell}
\usepackage{adjustbox}

\usepackage{flushend}
\setcopyright{ifaamas}  
\acmDOI{doi}  
\acmISBN{}  
\acmConference[   ]{Proc.\@ of the 19th International Conference on Autonomous Agents and Multiagent Systems (AAMAS 2020), B.~An, N.~Yorke-Smith, A.~El~Fallah~Seghrouchni, G.~Sukthankar (eds.)}{ }{ }  
\acmYear{2020}  
\copyrightyear{2020}  
\acmPrice{}  

\begin{document}

\title{Baconian: A Unified Open-source Framework for Model-Based Reinforcement Learning}  

\subtitle{Demonstration}



\author{Linsen Dong}

\affiliation{%
   \institution{School of Computer Science and Engineering, Nanyang Technological University}
   \city{Singapore} 
   \postcode{639798}
 }
 \email{linsen001@e.ntu.edu.sg}
 
 \author{Guanyu Gao}
 
 \affiliation{%
 	\institution{School of Computer Science and Engineering, Nanyang Technological University}
 	\city{Singapore} 
 	\postcode{639798}
 }
 \email{gygao@ntu.edu.sg}
 
 \author{Xinyi Zhang}
 
 \affiliation{%
 	\institution{School of Computer Science and Engineering, Nanyang Technological University}
 	\city{Singapore} 
 	\postcode{639798}
 }
 \email{zh0031yi@e.ntu.edu.sg}
 
 \author{Liangyu Chen}
 
 \affiliation{%
 	\institution{School of Electrical and Electronic Engineering, Nanyang Technological University}
 	\city{Singapore} 
 	\postcode{639798}
 }
 \email{lchen025@e.ntu.edu.sg}
 
 \author{Yonggang Wen}
 
 \affiliation{%
 	\institution{School of Computer Science and Engineering, Nanyang Technological University}
 	\city{Singapore} 
 	\postcode{639798}
 }
 \email{ygwen@ntu.edu.sg}

\begin{abstract}  
Model-Based Reinforcement Learning (MBRL) is one category of Reinforcement Learning (RL) algorithms which can improve sampling efficiency by modeling and approximating system dynamics. It has been widely adopted in the research of robotics, autonomous driving, etc. Despite its popularity, there still lacks some sophisticated and reusable open-source frameworks to facilitate MBRL research and experiments. To fill this gap, we develop a flexible and modularized framework, Baconian, which allows researchers to easily implement a MBRL testbed by customizing or building upon our provided modules and algorithms. Our framework can free users from re-implementing popular MBRL algorithms from scratch thus greatly save users' efforts on MBRL experiments.
\end{abstract}


\maketitle


\section{Introduction}
%
%
%
%
%
Model-Based Reinforcement Learning (MBRL) is proposed to reduce sample complexity introduced by model-free Deep Reinforcement Learning (DRL) algorithms \cite{nagabandi2018neural}.
Specifically, MBRL approximates the system dynamics with a parameterized model, which can be utilized for policy optimizing when the training data is very limited or costly to obtain in the real world.

%
%
Implementing a RL experiments from scratch can be tedious and bug-introducing.  
Luckily, many open-source frameworks have been developed to facilitate DRL research, including baselines \cite{baselines}, rllab \cite{duan2016benchmarking}, Coach \cite{coach}, and Horizon \cite{gauci2018horizon}. However, these frameworks are mainly implemented for model-free DRL methods, and lack enough support for MBRL.

Existing model-based frameworks are few and have some shortcomings. The work in \cite{DBLP:journals/corr/abs-1907-02057} gives a comprehensive benchmark over state-of-the-art MBRL algorithms, but the implementations are scattered across different codebases without a unified implementation, posing obstacles to conduct experiments with it. The work in \cite{finn2016guided} provides the implementations for Guided Policy Search(GPS) \cite{levine2016end}, which supports robotics controlling tasks. But it lacks support for other MBRL algorithms. Thus, a unified MBRL open-source framework is in need.
%
%
To fill this gap, we design and implement a unified MBRL framework, Baconian, by trading off the diversity of included MBRL algorithms against the complexity of the framework. 
Users can reproduce benchmark results or prototype their idea easily with it by a minimal amount of codes without understanding the detailed implementations. 
Moreover, the design of Baconian not only benefits the research of MBRL,  but is also applicable to other types of RL algorithms including model-free algorithms.
The codebase is available at \url{https://github.com/cap-ntu/baconian-project}. The demo video is available at \url{https://youtu.be/J6xI6qI3CvE}.
\begin{figure}
	\centering
	\includegraphics[width=0.85\columnwidth]{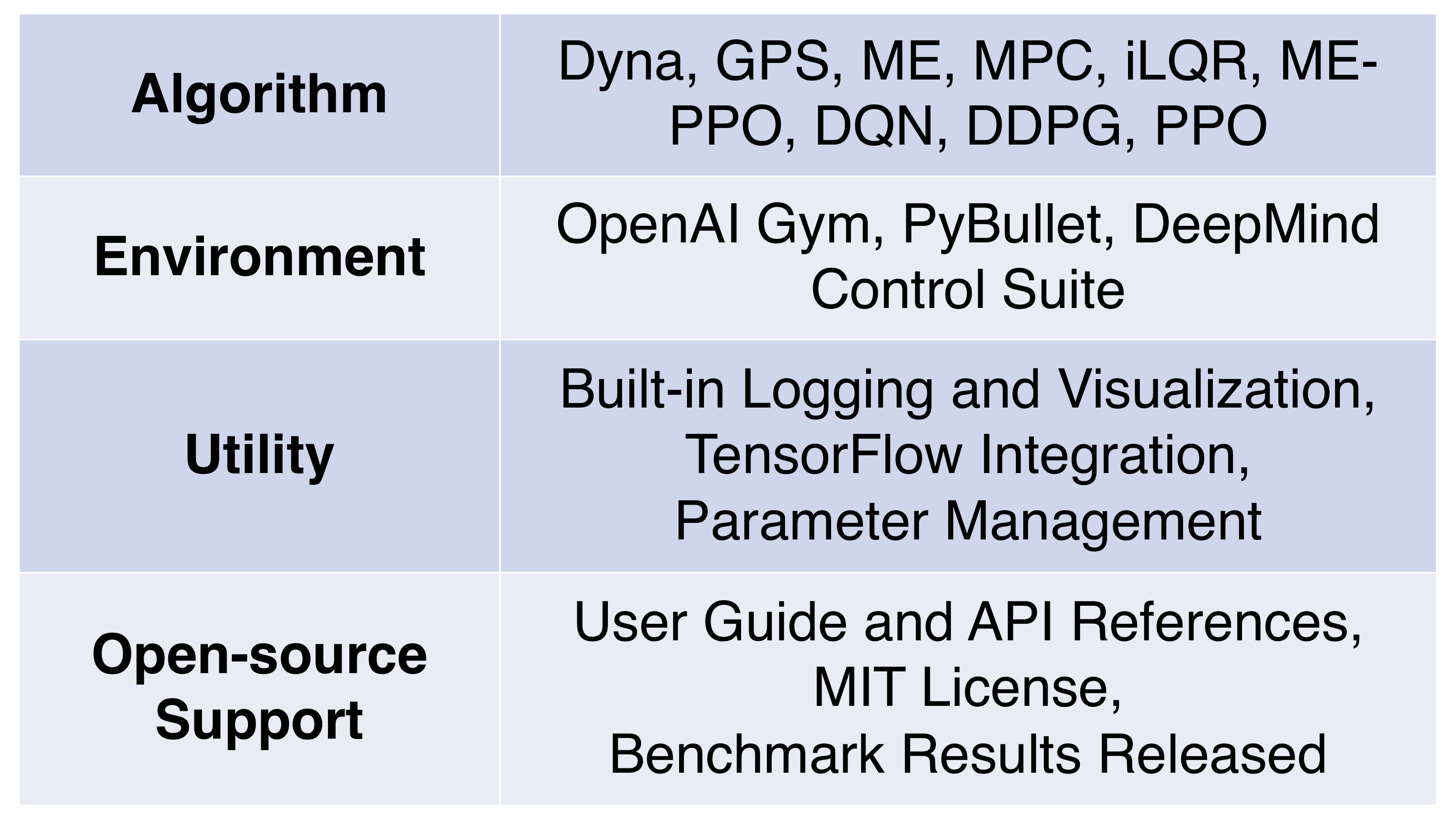}
	\caption{Feature list of Baconian.}
	\label{fig:feature}
\end{figure}
\section{Main Features}
Baconian supports many RL algorithms, test environments, and experiment utilities. We summarize the main features in Fig. \ref{fig:feature}.

\textbf{State-of-the-Art RL Algorithms}. We implement many widely used RL algorithms. For model-based algorithms, we implement Dyna\cite{sutton1991dyna}, ME-TRPO (Model-ensemble Trust Region Policy Optimization) \cite{kurutach2018model}, MPC (Model Predictive Control)\cite{richards2005robust}, iLQR (Iterative Linear Quadratic Regulator)\cite{tassa2012synthesis}, etc. Since many model-based algorithms are built upon model-free algorithms, we also implement some popular model-free algorithms including DQN\cite{DBLP:journals/corr/HosuR16}, DDPG\cite{DBLP:journals/corr/LillicrapHPHETS15}, and PPO\cite{DBLP:journals/corr/SchulmanWDRK17} in Baconian.

\textbf{Supported Test Environments}. To evaluate the performance of RL algorithms, it is a must to support a wide range of test environments. Baconian support OpenAI Gym\cite{gym}, RoboSchool\cite{klimov2017roboschool}, DeepMind Control Suite\cite{deepmindcontrolsuite2018}. These test environments cover most essential tasks in RL community.

\textbf{Experiment Utilities}. Baconian provides many utilities to reduce users' efforts on experiment set-up, hyper-parameter tuning, logging, result visualization, and algorithms diagnosis. We provide integration of TensorFlow to support neural network building, training, and managing. As the hyper-parameters play a critical role in RL experiments, we provide user-friendly parameter management utility to remove the tedious work of setting, loading, and saving these hyper-parameters.

\textbf{Open-source Support}. Baconian provide detailed user guides and API references\footnote{The documentation can be found at \url{https://baconian-public.readthedocs.io/en/latest/API.html}.}, so users can hand on Baconian easily and conduct novel MBRL research upon it. We also release some preliminary benchmark results in the codebase.
\section{Design and Implementation}
\begin{figure}
	\centering
	\includegraphics[width=1.0\columnwidth]{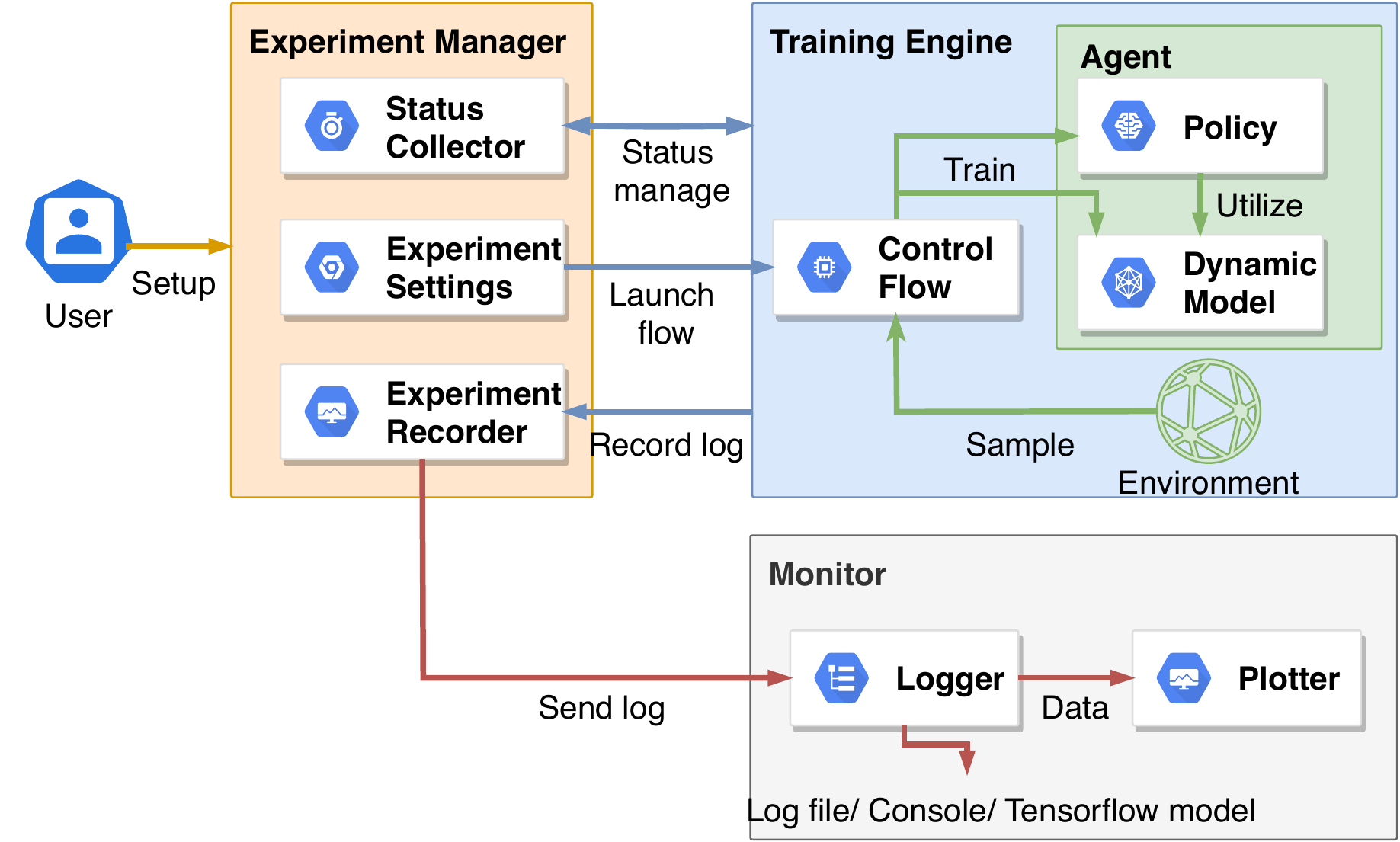}
	\caption{The system design of Baconian. The system is divided into three main modularized components to minimize the coupling for flexibility and maintainability.}
	\label{fig:system-diagram}
\end{figure}

Baconian consists of three major components, namely, Experiment Manager, Training Engine, and Monitor. 
The system overview of Baconian is shown in Fig. \ref{fig:system-diagram}. Various design patterns are applied to decouple the complicated dependencies across different modules to enable the easily extension and programming over the framework.

\subsection{Experiment Manager}
The Experiment Manager consists of Experiments Settings, Status Collector, and Experiment Recorder. Experiments Settings manages the creating and initialization of each module. Status Collector controls the status information collected across different modules to compose a globally shared status that can be used including learning rate decay, exploration strategy scheduling, etc. 
Experiment Recorder will record the information generated from the experiment, such as loss, rewards. Such information will be handed to the Monitoring layer for rendering or saving.

\subsection{Training Engine}
 Training Engine handles the training process of the MBRL algorithms. The novelty of the design lies in abstracting and encapsulating the training process as a Control Flow module, which controls the execution processes of the experiment based on the user's specifications, including the agent's sampling from environment, policy model and dynamics model optimization, and testing.
 MBRL algorithms can be complicated\cite{sutton1991dyna, nagabandi2018neural}. Such abstractions can decouple the tangled and complicated MBRL training processes into some independent tasks, which are further encapsulated as the sub-modules of Control Flow module for providing flexibility.
\subsection{Monitor}
Monitor is responsible for monitoring and recording of the experiment as it proceeds. This includes recording necessary loggings, printing information/warning/error, and rendering the results. 
\begin{figure}
	\centering
	\includegraphics[width=0.8\columnwidth]{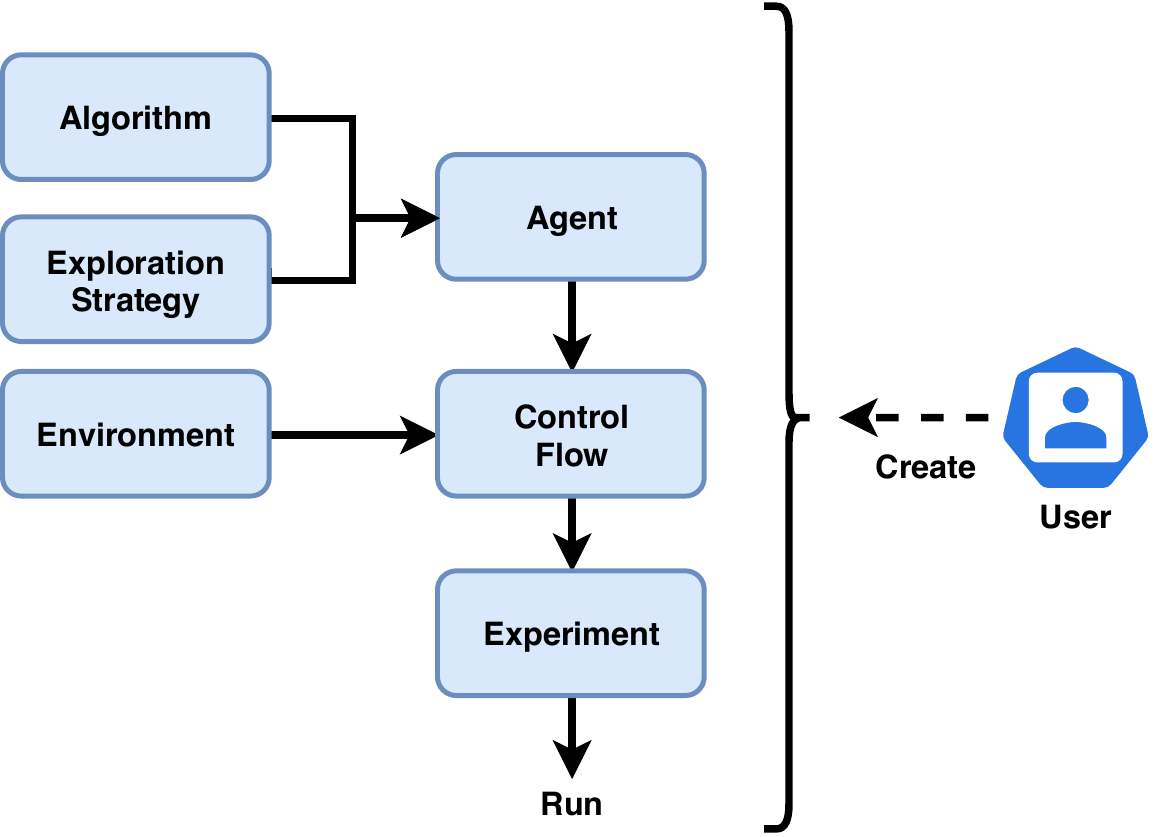}
	\caption{The procedure to create an MBRL experiment in Baconian. Each module is replaceable and configurable to reduce the effort of building from scratch.}
	\label{fig:process}
\end{figure}

\section{Usage}
This section presents the procedures to create a MBRL experiment with the essential modules in Baconian. The procedures are shown in Fig. \ref{fig:process}\footnote{For more details of how to configure these modules, please see the documentation page \url{https://baconian-public.readthedocs.io/en/latest/step_by_step.html} due to the page limit.}. For high flexibility, most of modules are customizable.
Meanwhile, user can directly adopt built-in benchmark module or codes if customization is unnecessary.

First, the user should create a environment and a RL algorithm module with necessary hyper-parameters configured, e.g., neural network size, learning rate. Algorithm module is usually composed of a policy module and a dynamics model module depending on different algorithms.
Then, the user needs to create an agent module by passing the algorithm module, and the exploration strategy module if needed, into it.

Second, the user should create a control flow module that defines how the experiments should be proceeded and the stopping conditions. This includes defining how much samples should be collected for training at each step, 
and what condition to indicate the completion of an experiment, etc. 
Some typical control flows have already been implemented in Baconian to meet the users' needs. 

Finally, an experiment module should be created by passing the agent, environment, control flow modules into it, and then launched. After that, the Baconian will handle the experiment running, monitoring, and results saving/logging, etc.

\section{Conclusion}
This paper presented a unified, reusable, and flexible framework, Baconian, for MBRL research.
%
%
It can reduce users' effort to conduct MBRL experiments and prototype new MBRL algorithms.
%
In the future, we will implement more state-of-the-art MBRL algorithms and benchmark them on different tasks.



\appendix
\section{List of Requirements}
To present the Baconian, we require a computer with operating system as Ubuntu 16.04/18.04, and installed with Python 3.5 or higher.

\bibliographystyle{ACM-Reference-Format}  
\bibliography{ref}  

\end{document}